\documentclass[10pt,twocolumn,letterpaper]{article}

\usepackage{iccv}
\usepackage{times}
\usepackage{epsfig}
\usepackage{graphicx}
\usepackage{amsmath}
\usepackage{amssymb}

\usepackage[breaklinks=true,bookmarks=false]{hyperref}

\iccvfinalcopy 

\def\iccvPaperID{****} 

\ificcvfinal\fi

\begin{document}

\title{The Second Place Solution for ICCV2021 VIPriors Instance Segmentation Challenge}

\author{Bo Yan, Fengliang Qi, Leilei Cao, Hongbin Wang\\
Ant Group\\
}

\maketitle
\ificcvfinal\thispagestyle{empty}\fi

\begin{abstract}

The Visual Inductive Priors(VIPriors) for Data-Efficient Computer Vision challenges ask competitors to train models from scratch in a data-deficient setting. In this paper, we introduce the technical details of our submission to the ICCV2021 VIPriors instance segmentation challenge. Firstly, we designed an effective data augmentation method to improve the problem of data-deficient. Secondly, we conducted some experiments to select a proper model and made some improvements for this task. Thirdly, we proposed an effective training strategy which can improve the performance. Experimental results demonstrate that our approach can achieve a competitive result on the test set. According to the competition rules, we do not use any external image or video data and pre-trained weights. The implementation details above are described in section 2 and section 3. Finally,  our approach can achieve 40.2\%AP@0.50:0.95 on the test set of ICCV2021 VIPriors instance segmentation challenge.

\end{abstract}

\section{Introduction}

Instance segmentation, a fundamental problem in computer vision, is widely used in image editing, image composition, autonomous driving, etc. Deep learning-based methods have achieved promising results for image instance segmentation over the past few years, such as Mask R-CNN~\cite{he2017mask}, PANet~\cite{liu2018path}, TensorMask~\cite{chen2019tensormask},  CenterMask~\cite{wang2020centermask}, SOLO~\cite{wang2020solo, wang2020solov2} . The main objective of the VIPriors instance segmentation challenge is to segment basketball players and balls on images recorded of a basketball court. Different from previous studies, VIPriors instance segmentation challenge does not allow using any pre-trained model, and training data is deficient.

\hfill

In order to address the problem of data-deficient, we designed an effective data augmentation method, which contains bbox-jitter, grid-mask, hue-transform, color-jitter, random-noise , etc. Then we conducted some experiments to verify whether previous studies are effective and made some improvements for this task. Finally, an effective training strategy is proposed to improve the model performance.

\hfill

\section{Approach}

Our approach mainly includes three parts: data augmentation, segmentation model, and training strategy. We introduce the proposed data augmentation strategy in Sec.2.1. The segmentation model is introduced in Sec.2.2. And the details of training strategy are introduced in Sec.2.3.

\subsection{Data Augmentation}

For VIPriors instance segmentation challenge, there are 184 images in the training set, 62 images in validation set, and 64 images in test set. The scenario seems uncomplicated since the task is only to segment basketball players and balls on images, 184 training images are insufficient. In order to generate enough images and make the model performs better and more robust, we designed an offline data augmentation strategy, including color-transform, quality-transform, filter-transform, and hue-transform.

There are four transforms for color-transform: random brightness, random color jitter, random saturation, and random sharpen. We randomly choose one of them as our color-transform to augment the image. There are also four transforms for quality-transform: random blur, random noise, random shuffle pixels, and random pixelization. We also randomly choose one of them as our quality-transform. PIL.ImageFilter is applied to filter-transform, including ImageFilter.DETAIL,  ImageFilter.EDGE\_ENHANCE, ImageFilter.SMOOTH, ImageFilter.MedianFilter, ImageFilter.ModeFilter, thus we randomly choose one of them as our filter-transform. Finally, we can change the hue of the image and get images with different colors by hue-transform. Followed by these four transforms, we can get an augmented image.

\begin{figure*}[ht]
	\centering
	\begin{tabular}
		{@{\hspace{0.0mm}}c@{\hspace{0.7mm}}c@{\hspace{0.7mm}}c@{\hspace{0.7mm}}c@{\hspace{0.7mm}}}
		\includegraphics[width=0.25\linewidth]{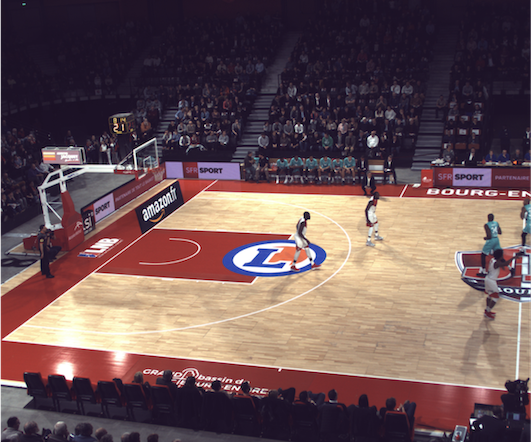}&
		\includegraphics[width=0.25\linewidth]{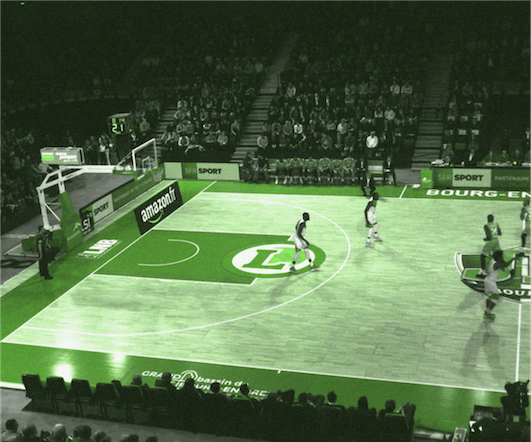}& 
		\includegraphics[width=0.25\linewidth]{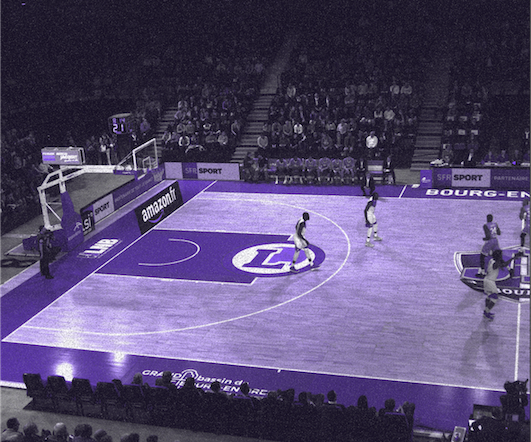}& 
		\includegraphics[width=0.25\linewidth]{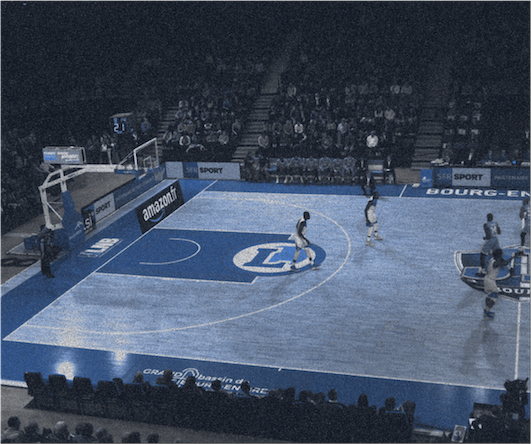}\\
		original image & augmented image1 & augmented image2 & augmented image3\\
	\end{tabular}
	\caption{Example augmented images by our proposed data augmentation strategy.}
	\label{Qualitative}
\end{figure*}

\begin{figure*}[ht]
	\centering
	\includegraphics[scale=0.465]{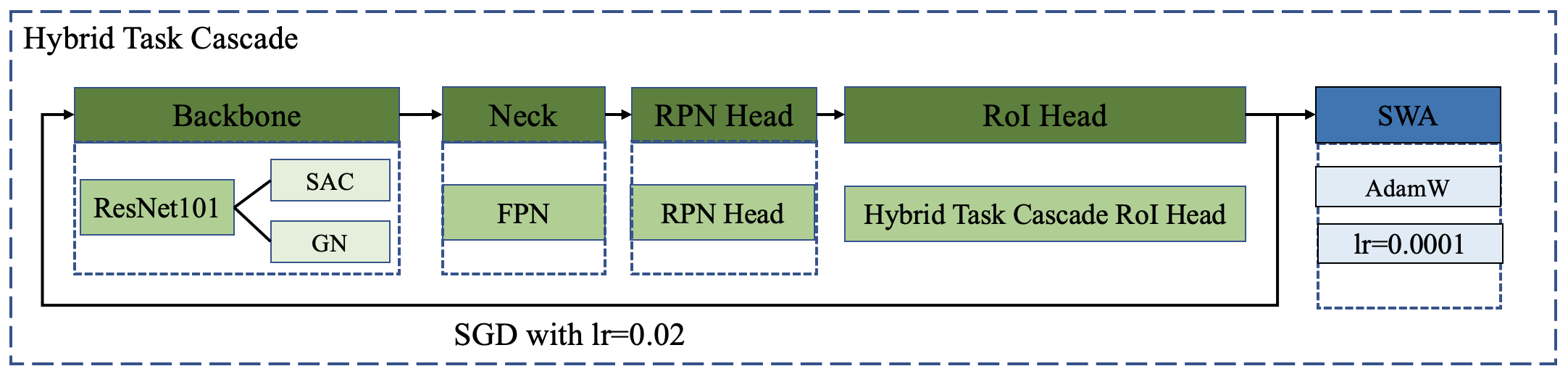}
	\caption{Our model architecture and training pipeline for VIPriors instance segmentation.}
	\label{pipeline}
\end{figure*}

We get ten augmented images for each image in training and validation set by the proposed offline augmentation strategy. Finally, our total training images are 2024, and total validation images are 682. Some example augmented images are shown in Figure 1. Some online data augmentation strategies, such as random-flip, random-crop,  bbox-jitter, grid-mask~\cite{chen2020gridmask} are also adopted during training.

\subsection{HTC-based Model}

Our instance segmentation model is based on Hybrid Task Cascade(HTC)~\cite{chen2019hybrid}. We use ResNet-101  as our backbone, but we convert the basic convolution layer in the backbone to switchable atrous convolution(SAC)~\cite{qiao2021detectors}, and also we use group normalization~\cite{wu2018group} to replace batch normalization for better performance.

\subsection{Training Strategy}

Firstly, to get a proper model for this task, we only use the training set to train the model and validate it on the validation set. Then we train the model with full training and validation set.  Stochastic Gradient Descent(SGD) with learning rate(lr)=0.02 is applied to this stage. Finally, SWA~\cite{zhang2020swa} training strategy is applied to finetune the model, which can make the model better and more robust. Adam with decoupled weight decay(AdamW)~\cite{loshchilov2018decoupled} with lr=0.0001 is applied during the SWA training stage.

Our model architecture and training pipeline are shown in Figure 2.

\hfill

\section{Experiments}

We train and evaluate our model from scratch without any pre-trained model, and we implement our experiments on a single GPU(NVIDIA Tesla V100 16GB).

\subsection{Training Details}

We only use the VIPriors instance segmentation training and validation set for training. The total images are 2706 include 2024 training images and 682 validation images after our proposed offline augmentation method. The images are randomly cropped to [1920, 1440], and some online augmentation methods are applied including random-flip, bbox-jitter, and grid-mask. The batch size for our training is set as 2.

\subsection{Experimental Results}

As shown in Table 1, our proposed approach finally achieves 40.2\%AP@0.50:0.95 on the VIPriors instance segmentation challenge test set.


\begin{table*}
\begin{center}
\begin{tabular}{lccc}
\hline
Methods & Training Data & Epochs & AP@0.50:0.95 \\
\hline
HTC-ResNet101-SAC-GN & training-2024 & 22 & 35.1\% \\
HTC-ResNet101-SAC-GN + SWA & training-2024 & 11 & 37.5\% \\
HTC-ResNet101-SAC-GN & training+validation-2706 & 20 & 36.5\% \\
HTC-ResNet101-SAC-GN + SWA & training+validation-2706 & 10 & 38.0\% \\
HTC-ResNet101-SAC-GN + SWA + Flip & training+validation-2706 & 10 & 38.7\% \\
HTC-ResNet101-SAC-GN + SWA & training+validation-2706 & 24 & 38.5\% \\
HTC-ResNet101-SAC-GN + SWA + TTA & training+validation-2706 & 24 & 40.2\% \\
\hline
\end{tabular}
\end{center}
\caption{Comparison results with HTC-ResNet101-SAC-GN on VIPriors instance segmentation test set}
\end{table*}

\subsection{Ablation Study}

This section elaborates on how we achieve the final result by ablation study to explain our approach. The baseline is HTC-ResNet101-SAC-GN. We train it only on the training set, in order to improve the recall of the model, soft nms~\cite{bodla2017soft} is used on the test stage for all experiments. The baseline achieves 35.1\% on the test set. Then we use SWA training strategy to finetune this model, after 11 epochs it achieves 37.5\% with a significant improvement of 2.4\%. Following experiments, we train the model on total images, including training and validation set. HTC-ResNet101-SAC-GN can achieve 36.5\%, the score increases to 38.0\% after SWA straining 10 epochs.  When adding the horizontal flip for the test set, it can achieve a better result  38.7\%. When adding test time augmentation(TTA), it can brings an improvement of 1.7\%. Our test time augmentation includes horizontal flip, multi-scale test with scale factors 0.8, 1.0, and 1.2.  Finally, we achieve 40.2\%  AP@0.50:0.95 on the test set of VIPriors instance segmentation.

\hfill

\section{Conclusion}

In this paper, we introduce the technical details of our submission to the ICCV2021 VIPriors instance segmentation challenge, including an effective data augmentation method, a proper model for this task, and an effective training strategy. Experimental results demonstrate that our approach can achieve a competitive result on the test set without external image or video data and pre-trained weights. Finally, our approach achieves 40.2\%AP@0.50:0.95 on the test set of VIPriors instance segmentation challenge.


{\small
\bibliographystyle{ieee_fullname}
\bibliography{egbib}
}

\end{document}